\documentclass[journal,twoside,web]{ieeecolor}
\usepackage{generic}
\usepackage{cite}
\usepackage{amsmath,amssymb,amsfonts}
\usepackage{multirow}
\usepackage{graphicx}
\usepackage{algorithm,algorithmic}
\usepackage{hyperref}
\hypersetup{hidelinks=true}
\usepackage{textcomp}

\def\BibTeX{{\rm B\kern-.05em{\sc i\kern-.025em b}\kern-.08em
    T\kern-.1667em\lower.7ex\hbox{E}\kern-.125emX}}
\begin{document}
\title{VERGE: Verification-Enhanced Refinement for Grounded Extraction of Early-Onset Colorectal Cancer Symptoms in Clinical Notes}
\author{%
\makebox[\textwidth][c]{Nikkie Hooman, Monarch Nigam, Amy E. Hughes, Rasmi G. Nair, and Mehak Gupta}%
\thanks{N. Hooman, M. Nigam, and M. Gupta are with the Department of
Computer Science, Southern Methodist University, Dallas, TX, USA
(e-mail: nikkieh@smu.edu; mnigam@smu.edu; mehakg@smu.edu).}%
\thanks{A. Hughes and R. Nair are with the Peter O'Donnell Jr. School of
Public Health, UT Southwestern Medical Center, Dallas, TX, USA
(e-mail: AmyE.Hughes@UTSouthwestern.edu;
rasmi.nair@utsouthwestern.edu).}%
\thanks{Corresponding author: Nikkie Hooman
(e-mail: nikkieh@smu.edu).}%
}
\maketitle

\begin{abstract}
Early-onset colorectal cancer is increasing among younger adults, yet red-flag symptoms in this age group have no evidence-based guidelines for follow-up testing, and structured encounter data do not capture the detail needed to support early detection and inform follow-up, including symptom duration, context, and family history, an established colorectal-cancer risk factor. This study aimed to develop and evaluate an automated method for extracting six red-flag symptoms and family-history risk status from free-text clinical notes. We developed VERGE, an agentic workflow in which an initial label and evidence are proposed using retrieval-augmented generation, then passed through a bounded verification-refinement cycle that checks textual grounding and clinical validity, corrects and re-checks a claim until resolved or a limit is reached, and escalates unresolved claims for human review. VERGE was evaluated on 4,033 clinician-labeled note-finding pairs against a single-agent baseline, a rule-based clinical language-processing baseline, and an alternative underlying language model. Compared with the single-agent baseline, VERGE reduced false positive findings, improving precision from 0.764 to 0.849 and MCC from 0.681 to 0.730, a balanced gain across the precision-recall trade-off, and resolved most flagged errors autonomously, with human review required for only 1.5 percent of claims. These results indicate that a bounded, verification-based workflow can reduce unnecessary positive findings without sacrificing the ability to detect true ones. This approach offers a path toward more reliable and trustworthy clinical language-processing tools to support colorectal cancer risk assessment in younger patients.
\end{abstract}


\section{Introduction}
\label{sec:introduction}

\IEEEPARstart{C}{olorectal} cancer is among the most commonly diagnosed cancers, with an estimated 158,850 new cases and 55,230 deaths projected in the United States in 2026 \cite{siegel2026colorectal}. While incidence has declined among adults aged 65 and older, it has risen by approximately 3\% per year among adults aged 20--49 \cite{siegel2026colorectal}. Early-onset colorectal cancer (EOCRC), diagnosed before 50 years of age, falls largely outside routine screening, which is recommended beginning at age 45 for average-risk adults \cite{wolf2026colorectal}. Diagnostic testing in this younger group is instead initiated only when red-flag (RF) symptoms, such as rectal bleeding, abdominal pain, and altered bowel habits, prompt clinical evaluation. These symptoms are common and non-specific, and no evidence-based guidelines govern which presentations warrant follow-up testing at what interval, contributing to diagnostic delays of approximately 4--6 months \cite{demb2024red}.

An RF symptom mentioned during a visit is not necessarily captured as an encounter-level diagnosis code until later in the diagnostic process, if at all, and even when a code is assigned, it does not capture the symptom-level detail needed to characterize presentation, support early detection, and inform follow-up. This detail, along with family-history information, an established risk factor for the disease \cite{wolf2026colorectal}, is more often available in free-text clinical notes \cite{lopez2025clinical,mowery2019determining}. Because RF symptoms are common in this age group yet infrequently attributable to colorectal cancer, reliably identifying patients who do not require further workup is as operationally important as identifying those who do. Extracting these symptoms from notes could therefore inform optimal triage, reducing unnecessary colonoscopy referrals while supporting timely diagnosis for patients who need it.

Recovering these findings from clinical notes is not straightforward. Evidence for a single symptom is often spread across a note rather than stated once, and these parts do not always agree: templated sections such as the review of systems (ROS) can deny a symptom even when other parts of the same note describe findings consistent with it, so relying on a single section risks missing evidence documented elsewhere. Symptoms may also be documented using varied wording, requiring interpretation beyond a keyword match, and a positive label must ultimately be supported by note text that directly documents the finding, with that evidence retained alongside the label to give end users a basis for trust. This makes EOCRC symptom abstraction more involved than concept detection alone, requiring a system to reconcile evidence across the note and determine whether it actually supports the claim.
 
We instantiate this as VERGE (\textbf{V}erification-\textbf{E}nhanced \textbf{R}efinement for \textbf{G}rounded \textbf{E}xtraction), a four-agent role-separated workflow for extracting six red-flag symptoms and one associated risk factor, family history of colorectal cancer, from individual clinical notes. Candidate labels and note-grounded supporting evidence are proposed, standardized into claims, and checked for textual grounding, clinical context, and target-specific support. VERGE autonomously corrects errors through a bounded Verifier--Refiner loop, in which an LLM judge decides whether a claim is adequately supported or must be flagged for refinement; unresolved claims are escalated for human review, increasing reliability and trustworthiness without requiring routine oversight of every claim. The bounded-loop output is retained as the VERGE prediction, and the complete evidence record accumulated through the agents, including loop state, exit status, and human-review status, is preserved to support downstream audit and clinical review.
 
This work makes four main contributions. First, we study evidence-verified extraction of EOCRC red-flag symptoms and family-history risk status at the level of individual note-level labels with explicit supporting evidence. Second, we design an Extractor that combines retrieval-augmented generation for contextual and terminology support with GRADE-informed evidence prioritization and a prompt-level rule that prioritizes narrative documentation over templated Review-of-Systems negations. Third, we introduce a bounded extract--compose--verify/refine workflow in which claims requiring correction are autonomously re-evaluated under fixed stopping criteria while preserving a complete audit trail. Fourth, we evaluate the system against baselines and development-stage comparisons, along with an analysis of verification and refinement.

\section{Related Work}
\label{sec:related_work}
 
Clinical NLP methods for extracting findings from colorectal cancer records have moved from rule-based systems to language models. Rule- and dictionary-based systems, such as MetaMap and cTAKES, map or link clinical text to standardized biomedical concepts, including concepts represented in the UMLS \cite{aronson2010,savova2010}. Context-sensitive methods further distinguish whether documented concepts are negated or occur under different contextual conditions: NegEx was developed for negation detection, while ConText extended this framework to attributes including temporality and experiencer \cite{chapman2001simple,harkema2009context}. Pretrained biomedical and clinical language models learn contextual representations directly from text; BioBERT and ClinicalBERT have demonstrated improvements across biomedical and clinical NLP tasks \cite{lee2020biobert,alsentzer2019publicly}. 

Within colorectal cancer specifically, prior work has used NLP and semantic analysis to characterize symptoms and symptom clusters from clinical records \cite{luo2021analyzing}. More recently, LLM-based approaches have been applied to colorectal cancer records, including the use of GPT-4o to extract symptoms and classify mode of presentation directly from clinical notes \cite{vetere2026mode}. 

Retrieval-augmented approaches can provide additional context for clinical extraction, although existing methods use retrieval in different ways. Bhattarai \textit{et al.} incorporated UMLS concepts into prompts for document-level clinical entity and relation extraction \cite{bhattarai2024document}, CLEAR retrieves clinically relevant portions of long clinical notes to support downstream LLM processing \cite{lopez2025clinical}, and MedCPT provides a biomedical information-retrieval model trained using large-scale PubMed search behavior \cite{jin2023medcpt}. These methods advance concept extraction and terminology coverage, but colorectal symptom-extraction work specifically has largely focused on producing the extracted label rather than verifying whether its supporting evidence is grounded in the source note, and retrieved biomedical context must remain distinct from the patient-specific evidence used to justify a label.

This gap in verification has motivated a shift toward multi-stage and multi-agent reasoning rather than a single generation. Systems such as MedAgents \cite{tang2024medagents} and MDAgents \cite{kim2024mdagents}, as well as multi-agent debate for clinical note error detection \cite{maiga2025error}, illustrate how role specialization and repeated checking can expose failures that a single generation pass would miss, consistent with a broader emphasis in clinical AI research on interpretable and auditable models that make the basis of their predictions explicit \cite{hooman2026multimodal}. VERGE builds on this direction, using role specialization and repeated verification within a bounded loop that preserves traceability and clinical oversight throughout autonomous correction.

\section{Dataset}
\label{sec:dataset}

We used existing data from an NIMHD-funded retrospective cohort study. The original dataset includes electronic health records (EHR) data for all primary care patients ages 18-49 years reporting at least one red flag symptom for colorectal cancer (abdominal pain, rectal bleeding, rectal pain, diarrhea, constipation, or weight loss) at Parkland Health between January 1, 2010 and December 31, 2020. Parkland Health (hereafter, Parkland) is an integrated safety-net health system providing care to more than 1 million uninsured residents of Dallas County, Texas through its 16 primary care clinics, specialty clinics, and tertiary-care hospital. Financial assistance is provided to all eligible Parkland patients based on a sliding scale. We excluded patients with a prior diagnosis of colorectal cancer and prior colectomy or gastrointestinal surgery. This study was approved by the UT Southwestern Medical Center
Institutional Review Board (IRB Protocol~\#~\mbox{STU-2024-0355}).

For the present pilot study, we used ICD-10 encounter diagnosis codes to stratify and sample, without replacement, 50 patient encounters for each of the six red-flag symptom categories (abdominal pain, rectal bleeding, rectal pain, constipation, diarrhea, and weight loss) from the 95,915 patients in the cohort, yielding 350 sampled encounters. A total of 583 clinical notes from 327 patients were available for annotation and were reviewed for the six red-flag symptoms and family history of colorectal cancer, yielding 4{,}081 possible note--finding pairs. A clinician reviewed each note to create gold-standard labels for each finding. After excluding 48 pairs with missing gold-standard labels, 4{,}033 labeled pairs remained, representing 578 notes from 326 patients. Table~\ref{tab:label_distribution} summarizes the gold-standard label distribution by target finding.

\begin{table}[!t]
\caption{Gold-standard label distribution by target finding.}
\label{tab:label_distribution}
\centering
\small
\renewcommand{\arraystretch}{1.08}
\setlength{\tabcolsep}{4pt}

\begin{tabular}{lrrrr}
\hline
\textbf{Symptom}
& \textbf{Samples}
& \textbf{Pos.}
& \textbf{Neg.}
& \textbf{Prev.} \\
\hline

Abdominal pain
& 576 & 243 & 333 & 42.2\% \\

Rectal bleeding
& 575 & 108 & 467 & 18.8\% \\

Rectal pain
& 575 & 77 & 498 & 13.4\% \\

Diarrhea
& 575 & 110 & 465 & 19.1\% \\

Constipation
& 577 & 79 & 498 & 13.7\% \\

Weight loss
& 577 & 67 & 510 & 11.6\% \\

Family history of CRC
& 578 & 18 & 560 & 3.1\% \\

\hline
\textbf{Total}
& \textbf{4{,}033}
& \textbf{702}
& \textbf{3{,}331}
& \textbf{17.4\%} \\
\hline
\end{tabular}
\end{table}

\section{Methodology}
\label{sec:methodology}

\begin{figure*}[!t]
\centering
\includegraphics[scale=0.34]{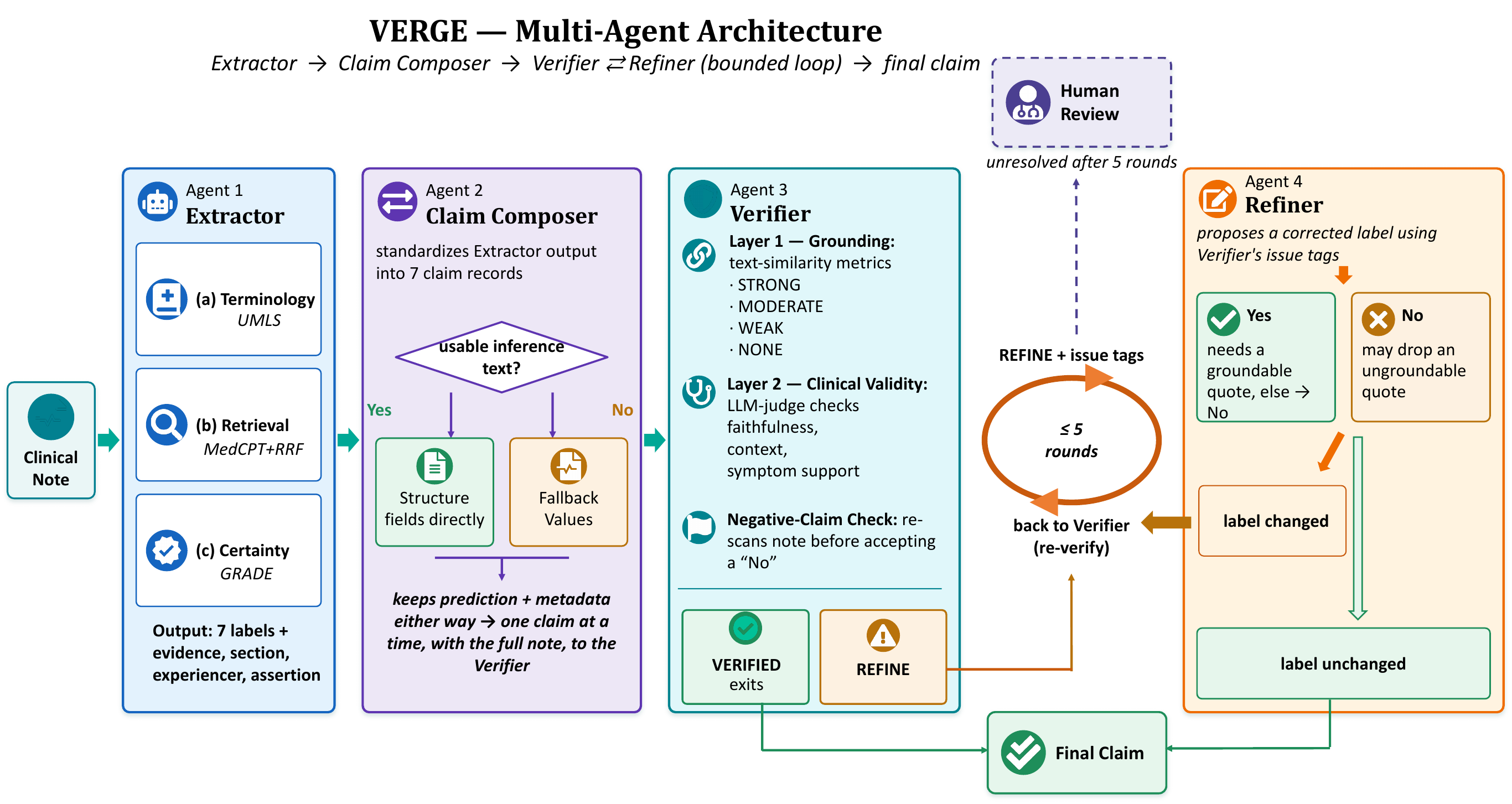}
\caption{Four-agent VERGE workflow. The Extractor proposes
evidence-linked labels, the Claim Composer standardizes claims, and the
bounded Verifier--Refiner loop verifies and revises them. Unresolved
claims are flagged for human review.}
\label{fig:verge_architecture}
\end{figure*}

\subsection{Problem Formulation and Framework Overview}
\label{sec:system_overview}

Let $\mathcal{N}=\{n_j\}_{j=1}^{N}$ denote the clinical notes and $\mathcal{F}=\{f_i\}_{i=1}^{7}$ denote the seven target findings: abdominal pain, rectal bleeding, rectal pain, diarrhea, constipation, weight loss, and family history of colorectal cancer. For each note--finding pair $(n_j,f_i)$, the gold-standard label is $y_{ij}\in\{0,1\}$, indicating whether the finding is documented as present in that note, and the goal is to predict $\hat{y}_{ij}\in\{\texttt{Yes},\texttt{No}\}$ with each positive prediction supported by evidence contained in $n_j$.

VERGE addresses this through a four-agent extract--compose--verify/refine workflow, shown in Fig.~\ref{fig:verge_architecture}: an Extractor proposes $\hat{y}_{ij}$ with note-linked evidence, a Claim Composer standardizes this into a claim, and a bounded Verifier--Refiner loop corrects and re-verifies the claim until it is resolved or flagged for human review, with the final loop state retained as VERGE's prediction. Each stage is described in detail below.

\subsection{Agent 1: Extractor}
\label{sec:extractor}
 
The Extractor produces a single joint prediction for all seven target findings from one clinical note in one language-model generation. Before generation, VERGE assembles biomedical context through three preparatory stages: terminology matching, retrieval, and evidence-quality prioritization.
 
\paragraph{Terminology matching.} For each of the seven target findings, a finding-specific dictionary is built by querying the UMLS Metathesaurus \cite{bodenreider2004} with the finding name to obtain its Concept Unique Identifier (CUI), then expanding that CUI through its UMLS synonym and related-concept relationships. This dictionary is combined with manually specified abbreviation matching. The note is then scanned against each finding's dictionary to identify the finding-relevant expressions.
 
\paragraph{Patient- and finding-conditioned retrieval.} Retrieval is conditioned jointly on the target finding and this note's content. For each finding, three MedCPT \cite{jin2023medcpt} queries are built: a base query that combines the finding with a general colorectal-cancer context, a note-anchored query derived from the aliases and UMLS synonyms detected in the note, and an ontology-expanded query that incorporates broader UMLS-derived terminology. All query variants are encoded with the MedCPT Query Encoder, while PubMed passage chunks are encoded with the MedCPT Article Encoder. For each query, the top 32 passages are retrieved by cosine similarity, and the resulting rankings are combined within each finding using Reciprocal Rank Fusion (RRF, $k=60$) \cite{cormack2009rrf}. Finding-specific phrase screening is applied to the fused ranking, with the unscreened ranking retained as a fallback. Up to 10 candidates per finding proceed to prioritization.
 
\paragraph{Certainty-based ranking.} Each retrieved passage inherits a GRADE certainty rating \cite{guyatt2008grade} from the body of evidence its source study belongs to, and this rating is used to rank the passage. For each finding, passages are ranked by their GRADE certainty, with RRF score breaking ties. At most two passages per finding are kept.
 
\paragraph{Final prompt construction.} The final prompt assembles the note, the finalized biomedical context (at most two passages per finding), and prompt-level rules that resolve conflicts between templated Review-of-Systems (ROS) entries and narrative sections such as the chief complaint and history of present illness, prioritizing narrative evidence over an ROS negation when the two disagree. This ROS conflict-resolution rule was developed in consultation with a practicing general internal medicine physician at Parkland. A positive prediction must be supported by evidence contained in the note, and this prompt is passed to the model in a single generation call.
 
\paragraph{Output schema.} For each finding, the Extractor returns a binary label, a confidence score (1--5), note-linked evidence, the note section it was drawn from, experiencer, and assertion status. Family history is produced in the same output but requires both an explicit biological relation and colorectal-, colon-, rectal-, or bowel-cancer specificity. The generated output is a flat JSON object that the Claim Composer parses and validates against this schema.

\subsection{Agent 2: Claim Composer}
\label{sec:composer}

The Claim Composer transforms the Extractor's joint output into seven standardized verifiable claim records, each holding the target finding, prediction, confidence, evidence, note section, experiencer, assertion status, and, where applicable, family-history fields. It validates this structure against the expected schema. When a finding's inference text is usable, it is parsed directly into these fields. When it is missing or malformed, a claim's missing structured fields still retain the Extractor's original inference text as its evidence. The original prediction and confidence are preserved. With no access to the clinical note, the Composer cannot introduce evidence or revise a label, keeping standardization a structural step rather than a second judgment pass. The seven resulting claims and the full note are passed individually to the Verifier.

\subsection{Agent 3: Verifier}
\label{sec:verifier}

The Verifier checks each standardized claim against the full clinical note for two distinct hallucination failure modes, and deterministically assigns either \texttt{VERIFIED} or \texttt{REFINE}.

\paragraph{Layer 1: Extrinsic hallucination (grounding).} This layer detects fabricated or unsupported evidence, cited text not actually derivable from the note, using exact substring matching, normalized exact matching, contiguous token-subsequence matching, ROUGE-L, modified BLEU, and BERTScore \cite{lin2004rouge,papineni2002bleu,zhang2020iclr-bertscore}. Normalization accounts for case, Unicode quotation marks, and whitespace. The contiguous token check requires at least four tokens; evidence with five or fewer non-whitespace characters is treated as unusable. BERTScore precision uses \texttt{roberta-large} without baseline rescaling, computed over overlapping chunks for notes exceeding the model's sequence limit, with the maximum precision retained. BLEU omits the standard brevity penalty, since the evidence quote is expected to be much shorter than the note. These measures combine into four grounding levels, \texttt{STRONG}, \texttt{MODERATE}, \texttt{WEAK}, and \texttt{NONE}, ordered from an exact match down to minimal correspondence; claims without usable evidence receive \texttt{NOT\_APPLICABLE}.

\paragraph{Layer 2: Intrinsic hallucination (clinical validity).} Textual grounding alone cannot establish clinical correctness. A span can be present in the note while its assertion and context do not support a positive interpretation, or it fails to directly support the requested finding. VERGE applies an LLM-as-judge to the full note to separately assess source faithfulness, valid assertion, context, correct experiencer and temporal progression, and direct support for the target finding. Family-history claims also require a biological relation and CRC specificity.

\paragraph{Negative claims.} A \texttt{No} prediction is not accepted automatically, since the absence of cited positive evidence does not by itself prove the finding is absent. Before a negative claim is finalized, the Verifier searches the note for text that could indicate the finding is actually present, using the same terminology and aliases used by the Extractor. When the search finds text mentioning a target finding, it checks whether the text affirmatively asserts that the finding is currently true for this patient. This rule filters out several common cases where the finding's name appears, but the text does not affirm it: a negated statement (e.g., ``denies rectal bleeding''), an old or resolved problem-list entry, and a mention that references the finding without confirming it occurred. A vague or loosely related match requires more specific corroboration, and family-history findings must also present a biological relation and the correct cancer type. If affirmative evidence that clears every check is found for the target finding, the negative claim is sent back for refinement. Together with Layer 1 and Layer 2, this gives VERGE a symmetric check against both false positives and false negatives.

\paragraph{Routing.} Grounding and clinical-validity results are mapped to \texttt{VERIFIED} or \texttt{REFINE} labels. A positive claim is \texttt{VERIFIED} only when grounding is \texttt{STRONG} or \texttt{MODERATE} and clinical validity explicitly passes, including relation and cancer-specificity checks for family history; otherwise it is \texttt{REFINE}. A negative claim is \texttt{VERIFIED} unless grounded affirmative evidence is recovered, in which case it becomes \texttt{REFINE}. A \texttt{VERIFIED} claim is stored as VERGE's prediction, and a \texttt{REFINE} claim is passed to the Refiner along with its issue tags and the full note.

\subsection{Agent 4: Refiner}
\label{sec:refiner}

The Refiner is invoked only on \texttt{REFINE} claims, re-examining the full note in light of the Verifier's issue tags to propose a corrected label. Every proposed correction is validated against the grounding and clinical-context validity before it replaces the current claim. A corrected \texttt{Yes} must include an exact or normalized note-grounded quotation or is converted to \texttt{No}. A corrected \texttt{No} may include an optional evidence quote but must include supporting text explaining why the earlier evidence was rejected. Family-history corrections to \texttt{Yes} additionally require a recognized biological relation and cancer-type specificity, and are converted to \texttt{No} if either is missing. The Refiner cannot exit the loop on its own: every corrected claim, whether or not its label changed, is returned to the Verifier for re-verification, since a correction can alter the supporting evidence even when the label is unchanged.

\subsection{Verifier-Refiner Loop} A claim whose label is unchanged from the one entering the \texttt{REFINE} round is label-stable and exits as the retained claim, while a claim whose label changes is returned to the Verifier for re-verification and, if flagged \texttt{REFINE} again, returns to the Refiner. This cycle is bounded to five rounds to allow repeated correction while preventing indefinite oscillation and limiting inference cost. A claim still unresolved after the final round retains its current state and is flagged for human review. A deterministic bounded controller orchestrates the loop, records the round-by-round trajectory, exit reason, and human-review status, and enforces the stopping and escalation rules.

\section{Experimental Design and Results}
\label{sec:experimental_design}

\subsection{Implementation Details}
\label{sec:implementation}

All language-model stages used \url{meta-llama/Meta-Llama-3.1-8B-Instruct} with greedy decoding. MedCPT retrieval and language-model inference were performed on one NVIDIA A100-SXM4-80GB GPU. The complete clinical note was preserved throughout extraction, and no clinical-note text was truncated. Clinical notes contained a median of 383 Llama tokens (IQR, 42--880; range, 2--5{,}110), while complete extraction prompts contained a median of 3{,}552 tokens (IQR,
3{,}098--4{,}004; maximum, 8{,}199). Further prompt design and model settings details are available in our \href{https://github.com/AI-for-Health-Data/SymptomDetection-EOCRC}{GitHub repository}.

\subsection{Performance Comparison Analysis}
\label{sec:comparisons_outcomes}

\begin{table}[!t]
\caption{Performance comparison across Extractor development stages, clinical NLP and single-agent baselines, and VERGE.}
\label{tab:overall_system_comparison}
\centering
\scriptsize
\renewcommand{\arraystretch}{1.15}
\setlength{\tabcolsep}{3pt}
\resizebox{\columnwidth}{!}{%
\begin{tabular}{@{}lrrrr@{}}
\hline
\textbf{Configuration}
& \textbf{Prec.}
& \textbf{Rec.}
& \(\boldsymbol{F_1}\)
& \textbf{MCC} \\
\hline
\textbf{Extractor Development} &&&& \\
Note-only   & 0.706 & 0.715          & 0.711 & 0.649 \\
ROS         & 0.625 & 0.755          & 0.684 & 0.614 \\
ROS + UMLS  & 0.620 & \textbf{0.758} & 0.682 & 0.611 \\
RAG         & 0.748 & 0.672          & 0.708 & 0.652 \\
\hline
\textbf{Single-Agent} &&&& \\
Extractor   & 0.764 & 0.705 & 0.733 & 0.681 \\
\hline
\textbf{Clinical NLP baseline} &&&& \\
\emph{medspaCy + ConText}
& 0.846 & 0.613 & 0.711 & 0.689 \\
\hline
\textbf{Complete VERGE} &&&& \\
\emph{Final} (Ministral)
& 0.861 & 0.693 & 0.767 & 0.723 \\
\emph{Final} (Llama)
& \textbf{0.849} & 0.702 & \textbf{0.769} & \textbf{0.730} \\
\hline
\end{tabular}%
}
\vspace{0.5mm}
\end{table}

We evaluate VERGE (Table~\ref{tab:overall_system_comparison}) through its developmental stages, against a single-agent baseline, an established clinical NLP baseline, and an alternative LLM backbone. All configurations are scored on the same 4{,}033 labeled pairs using precision, recall, $F_1$, and Matthews correlation coefficient (MCC), which summarizes classification performance using all four confusion-matrix outcomes.

The first category summarizes selected Extractor development configurations. \emph{Note-only} uses only the clinical note and a prompt to classify findings; \emph{ROS} resolves conflicts between templated Review-of-Systems negatives and narrative documentation; \emph{ROS + UMLS} adds UMLS-derived terminology \cite{bodenreider2004}; \emph{RAG} uses MedCPT dense retrieval \cite{jin2023medcpt} with within-finding Reciprocal Rank Fusion \cite{cormack2009rrf}; and \emph{Extractor} is the finalized configuration, adding GRADE-informed retrieval prioritization. As shown in Table~\ref{tab:overall_system_comparison}, adding ROS conflict handling raised recall by correcting false negatives caused by default ROS negation tags, which mark a finding as absent by template regardless of narrative evidence elsewhere in the note, but the same conflict resolution increased false positives, lowering precision. The finalized Extractor configuration corrected this trade-off, achieving the strongest balance of precision, recall, and MCC among all single-agent variants, and serving as the frozen input to VERGE.

As an established clinical NLP baseline, \emph{medspaCy + ConText} combines terminology matching with contextual assertion handling \cite{eyre2022launching,harkema2009context}, representing a non-agentic, rule-based approach to the same task. This baseline achieved high precision but comparatively low recall, reflecting a conservative operating point relative to the primary Llama-based VERGE.

\emph{Final} passes the frozen Extractor's output through the complete four-agent VERGE workflow, including the Claim Composer and bounded Verifier--Refiner loop, isolating the added effect of verification and refinement over the single-agent baseline. The primary Llama-based configuration increased precision from 0.764 to 0.849 and $F_1$ from 0.733 to 0.769 while maintaining similar recall (0.705 versus 0.702), achieving the highest $F_1$ and MCC among all evaluated configurations. This precision gain reflects a meaningful reduction in false positives, directly relevant to reliably identifying patients who do not require further workup, not only those who do. Three configurations, Note-only, ROS, and ROS + UMLS, in fact achieved higher recall than this configuration, with ROS + UMLS recovering the most true positive findings overall, but only by trading a recall reduction of roughly 2\% to 7\% relative to VERGE for a precision loss of 20\% to 37\%. Because red-flag symptoms in this age group are common but infrequently attributable to colorectal cancer, this trade-off works against the goal of triage: each additional true positive recovered by a higher-recall configuration comes bundled with substantially more false positives, each representing a patient unnecessarily flagged for further evaluation. VERGE's operating point favors the better-balanced trade-off, accepting a small recall cost for a much lower false-positive rate.

To assess whether this benefit depends on a specific underlying model, we repeat the Verifier--Refiner evaluation using \texttt{Ministral-3-14B-Instruct-2512-BF16} \cite{liu2026ministral} in place of Meta-Llama-3.1-8B-Instruct as the agent backbone. The workflow remained applicable with this second model family, indicating that the benefit of verification and refinement is not specific to a single backbone.

\subsection{Per-Finding Performance}
\label{sec:results_detailed}

Table~\ref{tab:per_symptom} reports per-finding performance for the Extractor and for VERGE, along with the number of label changes from Extractor to VERGE underlying each finding's performance shift. VERGE's largest gains occurred for rectal pain and rectal bleeding, driven almost entirely by improved precision with recall essentially unchanged, indicating that the Extractor was over-predicting positives for these findings, exactly the error pattern the Verifier's evidence checks are designed to catch. Family-history performance was volatile in both directions because its prevalence is much smaller than the other findings (Table~\ref{tab:label_distribution}), so a handful of label changes has an outsized effect.
 
Across all findings, VERGE's label changes were far more likely to correct a label (GS$+$) than to invalidate one (GS$-$), largely by removing false positives. This reflects VERGE's design: the main verification step targets unsupported positive claims, while missed positives rely on a narrower, dedicated search. Because these mechanisms are unequal in scope, corrections skew from positive to negative, a design effect rather than a bias against positive predictions.
 
\begin{table}[!t]
\caption{Per-finding performance and label-change outcomes from Extractor to VERGE. GS$+$ is a label change corrected to match the gold standard; GS$-$ is a change that invalidates a previously correct label.}
\label{tab:per_symptom}
\centering
\footnotesize
\renewcommand{\arraystretch}{1.05}
\setlength{\tabcolsep}{2.5pt}
\resizebox{\columnwidth}{!}{%
\begin{tabular}{lrrrrrrrr}
\hline
& \multicolumn{3}{c}{\textbf{Extractor}}
& \multicolumn{3}{c}{\textbf{VERGE}}
& \multicolumn{2}{c}{\textbf{Changes}} \\
\hline
\textbf{Finding}
& \textbf{P.} & \textbf{R.} & \(\boldsymbol{F_1}\)
& \textbf{P.} & \textbf{R.} & \(\boldsymbol{F_1}\)
& \textbf{GS$+$} & \textbf{GS$-$} \\
\hline
Abdominal pain
& 0.79 & 0.65 & 0.71
& 0.84 & 0.65 & 0.73
& 21 & 9 \\
Rectal bleeding
& 0.80 & 0.75 & 0.78
& 0.94 & 0.79 & 0.86
& 20 & 1 \\
Rectal pain
& 0.47 & 0.78 & 0.59
& 0.65 & 0.78 & 0.71
& 43 & 8 \\
Diarrhea
& 0.94 & 0.76 & 0.84
& 0.91 & 0.75 & 0.82
& 2 & 7 \\
Constipation
& 0.91 & 0.76 & 0.83
& 0.93 & 0.72 & 0.81
& 4 & 5 \\
Weight loss
& 0.89 & 0.72 & 0.79
& 0.87 & 0.70 & 0.78
& 1 & 3 \\
Family history of CRC
& 0.42 & 0.28 & 0.33
& 1.00 & 0.17 & 0.29
& 7 & 2 \\
\hline
\end{tabular}%
}
\end{table}

\subsection{Verification and Refinement Outcomes}
\label{sec:results_verification}

Table~\ref{tab:verification_analysis} summarizes the outcomes of the
367 claims (9.1\% of 4{,}033) that entered refinement. 80 claims (21.8\%) resolved with a changed label and 226 (61.6\%) resolved with the label unchanged, while 59 (16.1\%) remained unresolved because of label oscillation through the five-round bound and 2 (0.5\%) encountered refinement-stage operational errors, together accounting for the 61 claims (1.5\% of all processed claims) flagged for human review. The large majority of refined claims therefore resolved within the round limit, with only a small fraction reaching the five-round bound, suggesting that this limit functions primarily as a safeguard against rare non-convergence rather than as a constraint that regularly curtails correction before it can be completed.

\begin{table}[!t]
\caption{VERGE \texttt{VERIFIER} and \texttt{REFINER} outcomes.}
\label{tab:verification_analysis}
\centering
\small
\renewcommand{\arraystretch}{1.03}
\setlength{\tabcolsep}{2.5pt}

\begin{tabular}{@{}llrrr@{}}
\hline
\textbf{Stage} & \textbf{Outcome}
& \textbf{N} & \textbf{Total} & \textbf{\%} \\
\hline

\multirow{2}{*}{Verifier}
& Verified   & 3666 & 4033 & 90.9 \\
& Refine     & 367  & 4033 & 9.1 \\
\hline

\multirow{4}{*}{Refinement}
& Resolved with label flipped   & 80  & 367 & 21.8 \\
& Resolved with label unchanged & 226 & 367 & 61.6 \\
& Five-round unresolved         & 59  & 367 & 16.1 \\
& Refinement error              & 2   & 367 & 0.5 \\
\hline

\end{tabular}
\end{table}

\subsection{Nature of Verification Failures}
\label{sec:results_hallucination}

Table~\ref{tab:verifier_issues} summarizes the
issues among the 367 claims routed to refinement, split into Layer 1, extrinsic hallucination, and Layer 2, intrinsic hallucination. A claim can be flagged for more than one issue. Clinical-validity failures substantially outnumber grounding failures, showing the more common and consequential error is citing genuine note text for the wrong clinical claim, not fabricating evidence. A further share of flagged claims are negative claims where the Verifier--Refiner loop recovers evidence the Extractor missed, showing correction runs in both directions. Taken together, these findings support the paper's central motivation: lexical grounding checks alone would catch only a minority of the errors that matter, and a dedicated clinical-validity check is necessary to catch the more common and more subtle failure mode.

\begin{figure}[!t]
\centering
\includegraphics[
    scale=0.20,
        trim=1.5cm 3.5cm 1.4cm 3.5cm,
    clip
]{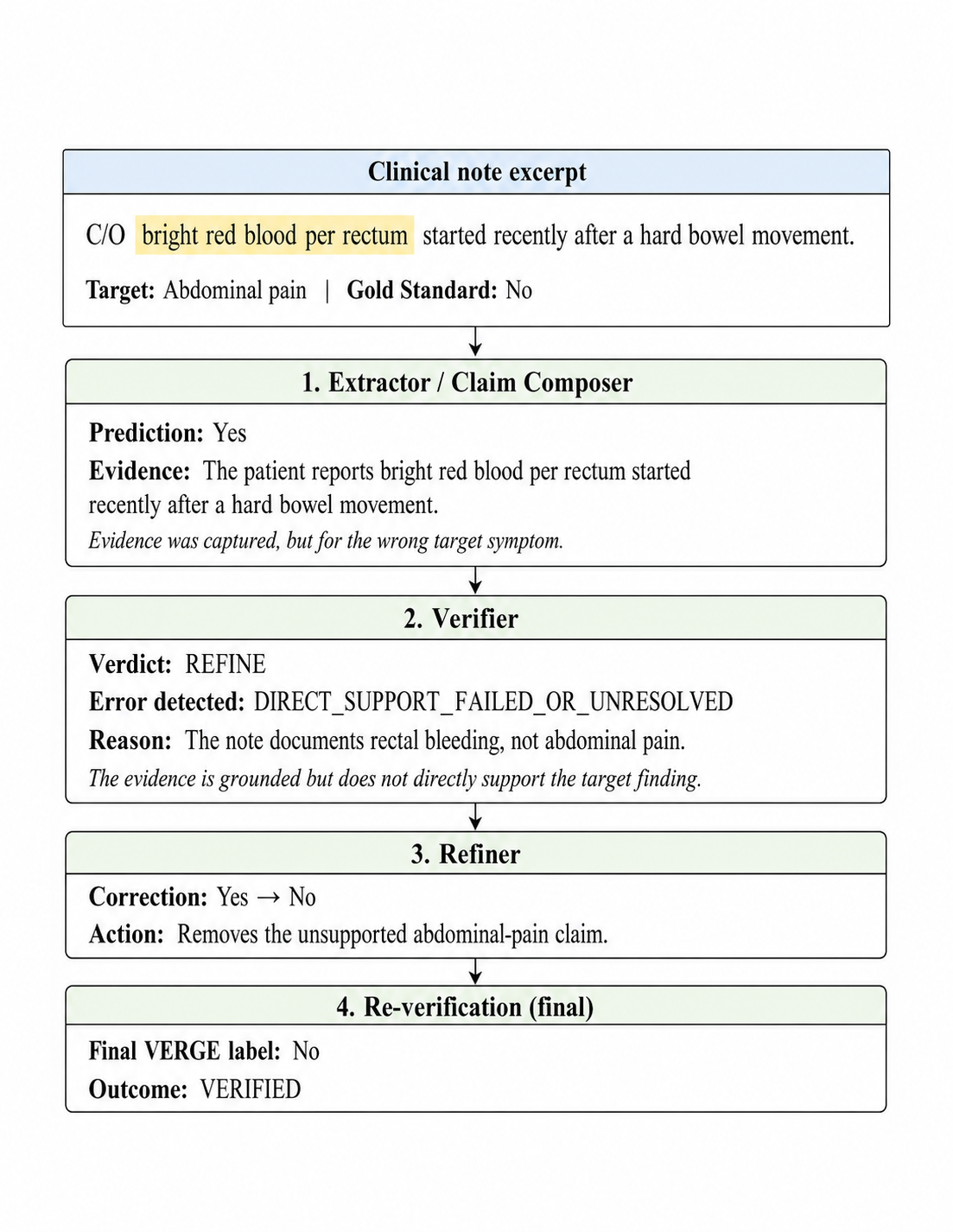}
\caption{Example of VERGE correcting a target-mismatch error through verification, refinement, and re-verification.}
\label{fig:example}
\end{figure}

Fig.~\ref{fig:example} illustrates this misattribution failure mode. For a note with abdominal pain as the target finding, the Extractor predicts \texttt{Yes} and cites genuine, correctly quoted evidence that the patient reports \emph{bright red blood per rectum started recently after a hard bowel movement}, evidence that is fully grounded but describes rectal bleeding, not abdominal pain. The Verifier flags this as a direct-support failure. The Refiner corrects the label to \texttt{No} and removes the unsupported claim, and re-verification confirms the correction as \texttt{VERIFIED}, matching the gold-standard label. The case shows why grounding alone is insufficient: this evidence would pass Layer 1 without difficulty, and only a dedicated clinical-validity check catches that it supports the wrong target finding.

\begin{table}[!t]
\caption{Initial Verifier issue profile among claims routed to refinement,
grouped by hallucination type.}
\label{tab:verifier_issues}
\centering
\small
\renewcommand{\arraystretch}{1}
\setlength{\tabcolsep}{6pt}

\begin{tabular}{@{}llrr@{}}
\hline
\textbf{Layer} & \textbf{Verification issue}
& \textbf{N}
& \textbf{\%} \\
\hline

\multirow{3}{*}{\shortstack[l]{Layer 1\\Extrinsic\\(grounding)}}
& Weak grounding      & 46 & 12.5 \\
& Missing evidence    & 13 & 3.5  \\
& Ungrounded evidence & 8  & 2.2  \\
\hline

\multirow{3}{*}{\shortstack[l]{Layer 2\\Intrinsic\\(clinical validity)}}
& Any clinical-validity failure & 222 & 60.5 \\
& CRC-specificity failure       & 7   & 1.9  \\
& Family-relation failure       & 3   & 0.8  \\
\hline

Negative claims
& Missed grounded & 189 & 51.5 \\
& affirmative evidence &  &  \\
\hline
\end{tabular}
\end{table}

\subsection{Exploratory Recovery of Missed Positive Claims via Generation-Time Uncertainty}
\label{sec:fn_recovery}

Negative claims present a distinct challenge for evidence-based correction, since they typically carry no cited evidence for the Verifier to assess. As an exploratory extension beyond the deterministic pipeline, we investigate whether generation-time uncertainty in the Extractor's original \texttt{Yes}/\texttt{No} decision can help identify negative claims that are more likely to be false negatives. Since language models do not output an explicit confidence score, uncertainty is recovered by comparing the model's generated output against constructed alternative continuations at their point of divergence, yielding a label preference and a binary entropy for that decision. Negative claims with elevated entropy are flagged as candidates and reviewed for missed evidence.  

Among 571 negative claims flagged by elevated entropy, 53 (9.3\%) were true false negatives, and 518 were true negatives, indicating that high entropy tracks genuinely ambiguous, difficult-to-interpret documentation rather than errors specifically. With targeted review, 5 of the 53 were correctly recovered as true positives, though this introduced 35 new false positives, raising recall slightly (0.702 to 0.709) while lowering precision, $F_1$, and MCC (to 0.802, 0.753, and 0.706). 

Qualitative review of the flagged false negatives identified two recurring patterns. The first involved resolving conflicting statements within narrative documentation, for example, a symptom described early in an encounter and later reported as resolved, distinct from the templated-versus-narrative conflict the Extractor already handles. The second reflected an opportunity for deeper recovery: the Verifier correctly rejected unsupported cited evidence, but correction did not always continue searching the note for valid alternative evidence before defaulting to a negative label.

\subsection{Statistical Comparison}
\label{sec:statistical_analysis}

To test whether VERGE's improvement over the Extractor is a reliable effect rather than sampling noise, we use a patient-cluster bootstrap where patients are resampled with replacement, keeping each patient's note--finding pairs together, and compute both systems' performance and take the difference between them. The VERGE-minus-Extractor difference is computed within each resample so that sampling variation cancels out. 

Table~\ref{tab:bootstrap_stats} reports each system's performance and this paired difference, each with a 95\% confidence interval. For precision, $F_1$, and MCC, the difference interval lies entirely above zero, indicating VERGE's improvement on these metrics is unlikely to be due to chance. For recall, the interval spans zero, consistent with no meaningful change between systems. These results support VERGE's gains as a dependable effect of verification and refinement rather than an artifact of this particular patient sample.

\begin{table}[!t]
\caption{Patient-cluster bootstrap estimates for the Extractor and VERGE, with the VERGE-minus-Extractor difference for each metric.}
\label{tab:bootstrap_stats}
\centering
\small
\renewcommand{\arraystretch}{1.15}
\setlength{\tabcolsep}{3pt}
\begin{tabular}{@{}llcc@{}}
\hline
\textbf{Metric} & \textbf{System}
& \textbf{Estimate}
& \textbf{Difference} \\
\hline
\multirow{2}{*}{Precision}
& Extractor & \shortstack{0.764\\(0.733--0.794)}
& \multirow{2}{*}{\shortstack{$+0.085$\\(0.057--0.112)}} \\
& VERGE & \shortstack{0.849\\(0.822--0.875)} & \\\hline
\multirow{2}{*}{Recall}
& Extractor & \shortstack{0.705\\(0.659--0.752)}
& \multirow{2}{*}{\shortstack{$-0.003$\\($-0.018$--0.013)}} \\
& VERGE & \shortstack{0.702\\(0.656--0.752)} & \\\hline
\multirow{2}{*}{$F_1$}
& Extractor & \shortstack{0.733\\(0.703--0.764)}
& \multirow{2}{*}{\shortstack{$+0.035$\\(0.020--0.051)}} \\
& VERGE & \shortstack{0.769\\(0.739--0.798)} & \\\hline
\multirow{2}{*}{MCC}
& Extractor & \shortstack{0.681\\(0.646--0.716)}
& \multirow{2}{*}{\shortstack{$+0.049$\\(0.030--0.069)}} \\
& VERGE & \shortstack{0.730\\(0.698--0.763)} & \\
\hline
\end{tabular}
\end{table}

\section{Discussion}
\label{sec:discussion}

Across development, baseline, and cross-model comparisons, VERGE's central result shows that it achieves a better-balanced precision-recall trade-off than other evaluated alternative, substantially reducing unsupported positive predictions. In practice, this means VERGE becomes considerably more trustworthy when it flags a finding as present, without becoming meaningfully less able to catch findings that are genuinely there. This behavior is clinically meaningful precisely because it serves triage rather than detection alone: a false positive in this setting corresponds to a patient who would be unnecessarily flagged for further evaluation, so reducing false positives without sacrificing sensitivity means fewer patients face unwarranted follow-up while patients who truly need it continue to be identified at essentially the same rate.
 
The dominant error VERGE caught was misattribution, not fabrication: real, well-quoted evidence applied to the wrong finding. This is a more insidious failure than an invented citation, since it looks trustworthy on the surface while still being clinically wrong. This has a broader implication for how clinical AI systems
should be evaluated: a system judged only on whether its
citations are real, without checking whether those citations
are relevant to the specific claim they support, could appear
highly trustworthy by standard grounding metrics while still
making clinically consequential errors at scale.
 
Generation-time uncertainty tracked genuine ambiguity in the source text rather than a specific error type, making it too imprecise to drive automated correction alone but well suited to prioritizing claims for secondary review. This suggests that deeper reasoning over a note's internal structure is a promising direction for future work. Such a direction would move beyond isolated claim checks toward managing conflicts through least-to-most reasoning consolidation and continuing to search for valid support once an initial citation is rejected.These extensions would require careful architectural modification to the Refiner and Verifier, guided by clinical collaboration.
 
These results reflect retrospective note abstraction, not validation as an autonomous screening tool. The bounded verification loop and the grounding/validity separation may generalize to other clinical extraction tasks, but the finding list, terminology, and evidence rules are EOCRC-specific. Labels came from a single clinician, and family history relied on only 18 positive cases, limiting confidence in that specific result. Future work could extend both labeling and model adjudication to reflect multi-rater consensus. These caveats aside, the overall pattern supports bounded, role-separated verification as a workable design for keeping each label's evidence verifiable.

\section{Conclusion}
\label{sec:conclusion}

We presented VERGE, an agentic framework that pairs bounded, autonomous verification with an explicit, auditable evidence record for extracting red-flag symptoms and family-history risk status from EOCRC clinical notes. Across development, baseline, and cross-model comparisons, VERGE achieved the best precision-recall balance of any evaluated configuration, without a corresponding loss in sensitivity, a pattern directly relevant to reliably identifying which patients do and do not warrant further workup. Our error analysis further shows that the dominant failure in this setting is misattributed, not fabricated, evidence, reinforcing the need for verification that checks clinical validity alongside textual grounding.

This work was conducted on a single-institution retrospective cohort with gold-standard labels from one clinician. We view VERGE as a step toward clinical information-extraction systems whose autonomy is exercised within human-defined bounds, keeping every label's supporting evidence available for clinical review rather than treated as a byproduct.

\section*{References}
\bibliographystyle{IEEEtran}
\bibliography{ref}

\end{document}